\documentclass[journal]{IEEEtran}
\usepackage{latexsym}
\usepackage{graphicx}
\usepackage{amsfonts,amssymb,amsmath}
\usepackage{hyperref}
\def\BibTeX{{\rm B\kern-.05em{\sc i\kern-.025em b}\kern-.08em T\kern-.1667em\lower.7ex\hbox{E}\kern-.125emX}}

\usepackage{algorithm}
\usepackage{algorithmic}

\usepackage{url}
\newfont{\bbb}{msbm10 scaled 500}

\newfont{\bb}{msbm10 scaled 1100}

\newcommand{\hv}{{\bf h}}

\newcommand{\pv}{{\bf p}}
\newcommand{\qv}{{\bf q}}

\newcommand{\wv}{{\bf w}}
\newcommand{\vv}{{\bf v}}
\newcommand{\xv}{{\bf x}}
\newcommand{\yv}{{\bf y}}

\newcommand{\Bm}{{\bf B}}

\newcommand{\Gm}{{\bf G}}

\newcommand{\Ym}{{\bf Y}}

\renewcommand{\arg}{{\hbox{arg}}}

\usepackage{bbm}
\usepackage{subcaption}
\usepackage{tabularx}

\usepackage{multirow}
\usepackage{verbatim}
\usepackage{color}
\usepackage{mathrsfs}
\usepackage{float}
\usepackage{bm}
\usepackage{booktabs}
\usepackage{diagbox}

\include{defn}

\def\argmin{\operatornamewithlimits{arg\,min}}

\newcommand{\beqa}{\begin{eqnarray}}
\newcommand{\eeqa}{\end{eqnarray}}

\begin{document}

\title{A Meta-Learning Approach to Compute Optimal Power Flow Under Topology Reconfigurations}

\author{Yexiang Chen, Subhash Lakshminarayana~\IEEEmembership{Senior Member, IEEE}, Carsten Maple, and H. Vincent Poor~\IEEEmembership{Fellow, IEEE}
\thanks{ Yexiang Chen and Subhash Lakshminarayana are with School of Engineering, University of Warwick, UK (e-mail: \{yexiang.chen, subhash.lakshminarayana\}@warwick.ac.uk). Carsten Maple is with Warwick Manufacturing Group, University of Warwick, UK (e-mail: CM@warwick.ac.uk). H. Vincent Poor is with Department of Electrical and Computer Engineering, Princeton University, Princeton, NJ 08544, USA (e-mail: poor@princeton.edu).}}


\maketitle
\begin{abstract}
Recently there has been a surge of interest in adopting deep neural networks (DNNs) for solving the optimal power flow (OPF) problem in power systems. Computing optimal generation dispatch decisions using a trained DNN takes significantly less time when compared to  conventional optimization solvers. However, a major drawback of existing work is that the machine learning models are trained for a specific system topology. Hence, the DNN predictions are only useful as long as the system topology remains unchanged. Changes to the system topology (initiated by the system operator) would require retraining the DNN, which incurs significant training overhead and requires an extensive amount of training data (corresponding to the new system topology). To overcome this drawback, we propose a DNN-based OPF predictor that is trained using a meta-learning (MTL) approach. The key idea behind this approach is to find a common initialization vector that enables fast training for any system topology. The developed OPF-predictor is validated through simulations using benchmark IEEE bus systems. The results show that the MTL approach achieves significant training speeds-ups and requires only a few gradient steps with a few data samples to achieve high OPF prediction accuracy and outperforms other pretraining techniques.
\end{abstract}


\section{INTRODUCTION}

\IEEEPARstart{T}{he} optimal power flow (OPF) problem involves the computation of minimum cost generation dispatch subject to the power flow equations and the grid's operational constraints (e.g., voltage/power flow limits, etc.). Power grid operators must solve the OPF problem repeatedly several times a day in order to ensure economical operation. The OPF problem under the generalized alternating current (AC) power flow model is non-convex, and solving them using conventional optimization solvers can be computationally expensive. The growing integration of renewable energy and the power demand uncertainty necessitates solving the OPF problem repeatedly at a significantly faster time scale (in the order to seconds) to respond to the changing system states, leading to significant computational challenges  \cite{Yang2017}.

To overcome this challenge, there has been a significant interest in adopting machine learning (ML) techniques to speed up the computation of the OPF problem. The ML models can be trained \emph{offline}, and the trained model can be used \emph{online} to support the computation of the optimal generation dispatch. The main advantage of this approach is that online computations are cheap, and hence, they can speed up OPF computation significantly. ML has been applied in a number of different ways to support OPF computation.


The most straightforward approach is to use ML models (e.g., DNNs) to directly learn the mapping from the load inputs to the OPF outputs \cite{Sun2018, Canyasse2017, PanDeepOPF2019, Zamzam2020}. The real-time load demands are fed as inputs to the trained ML model, and the corresponding OPF solution are computed as outputs. As compared to using conventional optimization solvers, this end-to-end approach was shown to provide up to $100$ times speed-up for the DC OPF problem \cite{PanDeepOPF2019} and $20$ times speed-up for the AC OPF problem \cite{Zamzam2020}. This direct approach effectively eliminates the time requirement for conventional optimization solvers in real-time. Furthermore, ML under the direct approach has been used in other applications such as 
scheduling under outages \cite{Dalal2019} and to provide decision support for distributed energy resources \cite{Karagiannopoulos2019, Dobbe2020}.

Different from this approach, ML can also be used indirectly to speed up conventional optimization solvers. For example, ML can be used to learn the set of active constraints at optimality; this approach was used to solve the direct current (DC) OPF problem in \cite{NgDCOPT2018, Deka2019, Jahanbani2018, Baker2018} and  inactive constrains in \cite{Pineda2020}.
Alternatively, ML can also be used to compute the so-called \emph{warm} start points for optimization solvers, an approach that is especially useful to solve the non-linear AC OPF problem \cite{Baker2019, JameiMeta2019}. Compared to these indirect approaches \cite{NgDCOPT2018}-\cite{JameiMeta2019}, the direct approach can achieve greater computational speed-up.

Recent works \cite{PanXiang2021, VenzkeNew2020, Venzke2021} have also provided feasibility guarantees, i.e., ensured that the solutions proposed by the ML models satisfy the power grid's operational constraints (e.g., line/voltage limits, etc.). Specifically, \cite{PanXiang2021} proposes a method to ensure feasibility by calibrating the power grid physical limits used in the DNN training. The worst-case guarantees with respect to physical constraint violations for the DNN's OPF solution were derived in \cite{VenzkeNew2020,Venzke2021} and used these results to reduce the worst-case errors.

However, a major drawback of existing work \cite{Sun2018}-\cite{Venzke2021} is that they are designed for a specific system configuration. As such, they remain effective only as long as the system topology remains fixed. Nevertheless, topology reconfigurations by transmission switching and impedance changes are essential parts of grid operations that can improve the grid's performance from both operational efficiency and reliability point of view \cite{Hedman2011, KoradTPS2013, RogersDFACTS2008}. These measures have gained increasing attention recently. For instance, perturbation of transmission line reactances (using distributed flexible alternating-current transmission systems, D-FACTS devices \cite{DFACTS2007}) is finding increasing applications in power flow control to minimize the transmission power losses \cite{RogersDFACTS2008} and cyber defense \cite{LakshDSN2018, LiuMTD2018, LakshCCPA2019}. Similarly, grid operators also perform transmission switching and topology control to ensure economic and reliable system operations \cite{Hedman2011, KoradTPS2013}.

Active topology control poses significant challenges in the use of DNNs for OPF prediction. A DNN trained under a specific system configuration might not be able to provide correct OPF outputs under a different system configuration. This is because the mapping between the load inputs and the OPF outputs will change due to the changes in the system topology. Indeed, our results show that DNNs trained on a specific topology have a poor generalization performance when the system topology changes. Complete retraining with the new system configuration will require significant amounts of training data and time, thus negating the computational speed-up achieved by DNN prediction.

To address these shortcomings, we propose a novel approach in which we train the DNN-based OPF predictor using a meta-learning (MTL) approach. The main idea behind MTL is to a find good initialization point that enables fast retraining for different system configurations.  Specifically, we use the so-called model-agnostic MTL approach \cite{FinnMAML2017}, which finds the initialization point in such a way that a few gradient steps with a few training samples from any system configuration will lead to good prediction performance. Thus the method is well suited to predict OPF solution under planned topology re-configurations. To the best of our knowledge, this work is the first to utilize MTL in a power grid context.

We conduct extensive simulations using benchmark IEEE bus systems. We compare the performance of MTL against several other approaches. They include (i) ``Learn from scratch'': in which, there is no pretraining, i.e., when the system is reconfigured, we initialize the DNN weights randomly and train them using the OPF data from the new system reconfiguration. (ii) ``Learn from a joint training model'': in which, during the offline phase, we train a DNN model from a combined dataset consisting of OPF data from several different topology configurations. Then during the online phase, we initialize the weights on the DNN using this model and fine-tune it using OPF data from the new system configuration. (iii) ``Learn from the closet model'': in which, during the offline phase, we train several DNN models separately using OPF datasets from different topology configurations (i.e., one DNN for each system configuration). Then, during the online phase, when the topology is reconfigured, we choose the model that achieves the best prediction performance on the new configuration and choose its weight as the initial DNN's weights. The weights are then fine-tuned using OPF data from the new configuration.

We verify the efficacy of the proposed approach by simulations conducted using IEEE bus systems. We generate the OPF data using the MATPOWER simulator and implement the ML models using Pytorch.  The results show that the proposed MTL approach can achieve significant training speed-ups and achieve high accuracy in predicting the OPF outputs. For instance, for the IEEE-118 bus system, MTL can achieve greater than $99 \%$ OPF generation prediction accuracy for a new system configuration with less than $10$ gradient updates and $50$ training samples. Furthermore, MTL can achieve a much higher prediction accuracy as compared to complete retraining (i.e., training from scratch), especially in the limited data regime (i.e., when the number of training data samples from a new system configuration are limited). MTL also outperforms the other two pretraining methods in terms of the OPF prediction accuracy and takes significantly less time/storage in the pretraining phase. Thus the method is well suited to predict the OPF solution under planned topology reconfigurations.

The rest of this paper is organized as follows. Section~2 introduces the power grid model, OPF problem and DNN approach. Section~3 details the proposed MTL method. Section~4 presents the simulation results to show the effectiveness of MTL over other pretraining methods and the conclusions are presented in Section~5. Some additional simulation results are included in Appendix.


\section{Preliminaries}
\subsection{Power Grid Model}
We consider a power grid with $\mathcal{N} = \{0, 1, 2, \ldots,N\}$ buses. A subset of the buses $\mathcal{G} \subseteq \mathcal{N}$ are equipped with generators. Without the loss of generality, we assume bus~0 to be the slack bus whose voltage is set to $1.0 \angle 0$ pu.
Since the interest of this paper is grid topology reconfigurations, we consider $M$ different grid topologies, where each topology differs with respect to the bus-branch connectivity and transmission line impedances. We assume that the nodes of the power grid always remain connected (among all the considered topologies).  
We let $\mathcal{L}^{(m)} = \{1, 2, \ldots, L^{(m)}\}$ denote the set of transmission lines under topology $m \in \{1, 2, \ldots,M\}.$ Further, we let $\Ym^{(m)} = \Gm^{(m)} + j \Bm^{(m)}$ denote the bus admittance matrix under topology $m,$ where $\Gm^{(m)}$ and $\Bm^{(m)}$ denote conductance and susceptance respectively \cite{wood1996power}.

Under topology $m$, let $P^{(m)}_{G_i}$ ($P^{(m)}_{D_i}$) and $Q^{(m)}_{G_i}$ ($Q^{(m)}_{D_i}$) denote the active and reactive power generations (demands) at node $i \in \mathcal{N}$ respectively. 
The complex voltage at node $i\in \mathcal{N}$ under topology $m$ is denoted by $V^{(m)}_i = |V^{(m)}_i| \angle \theta^{(m)}_i,$ where $|V^{(m)}_i|$ is the voltage magnitude and $\theta^{(m)}_i$ is the voltage phase angle. According to the AC power flow model, these quantities are related as
\begin{align}
P^{(m)}_{G_i} - P^{(m)}_{D_i} &= \vert V^{(m)}_i \vert \sum_{ j \in \mathcal{N}}    V^{(m)}_j  (G^{(m)}_{i,j}cos(\theta^{(m)}_{i,j}) \nonumber \\ & \qquad \qquad \qquad \qquad  +B^{(m)}_{i,j}sin(\theta^{(m)}_{i,j})), \label{eqn:RealPF} \\
Q^{(m)}_{G_i} - Q^{(m)}_{D_i} &= \vert V^{(m)}_i \vert \sum_{ j \in \mathcal{N}}     V^{(m)}_j  (G^{(m)}_{i,j}sin(\theta^{(m)}_{i,j}) \nonumber \\ & \qquad \qquad \qquad \qquad  -B^{(m)}_{i,j}cos(\theta^{(m)}_{i,j})), \label{eqn:ReacPF}
\end{align}
where $\theta^{(m)}_{i,j} = \theta^{(m)}_i-\theta^{(m)}_j.$

\emph{Optimal Power Flow Problem:} The OPF problem computes the minimum cost generation dispatch for a given load condition constrained to the power flow equations and power generation/voltage constraints. Mathematically, the OPF problem can be stated as follows:
\begin{align}
\min_{ \substack{P^{(m)}_{G}, \\ Q^{(m)}_{G}, V^{(m)}}} &  \sum_{i \in \mathcal{G}} C_i (P^{(m)}_{G_{i}})  \label{eqn:OPF_AC} \\
s.t. & \ \eqref{eqn:RealPF}, \ \eqref{eqn:ReacPF}, \nonumber\\
&{P^{\min}_{G_{i}}} \leq P^{(m)}_{G_{i}} \leq {P^{\max}_{G_{i}}}, \forall i \in \mathcal{G} \label{eqn:constrain_P}\\
 &{Q^{\min}_{G_{i}}} \leq Q^{(m)}_{G_{i}} \leq {Q^{\max}_{G_{i}}},
\forall i \in \mathcal{G} \label{eqn:constrain_Q}\\
&{V}^{\min}_i \leq V^{(m)}_{i} \leq {V}^{\max}_i, \forall i \in \mathcal{N} \label{eqn:constrain_V},
\end{align}
where $C_i(\cdot)$ is the generation cost 
at bus $i \in \mathcal{G}$. Further, ${P}_{G_i}^{\max}$ (${P}_{G_i}^{\min}$), ${Q}_{G_i}^{\max}$ (${Q}_{G_i}^{\min}$) and ${V}^{\max}_i$ (${V}^{\min}_i$) denote the maximum (minimum) real/reactive power generations and nodal voltage limits at node $i$ respectively.

\subsection{DNN Approach for the OPF problem}

\begin{figure}[!t]
	\centering
	\includegraphics[width=0.4\textwidth]{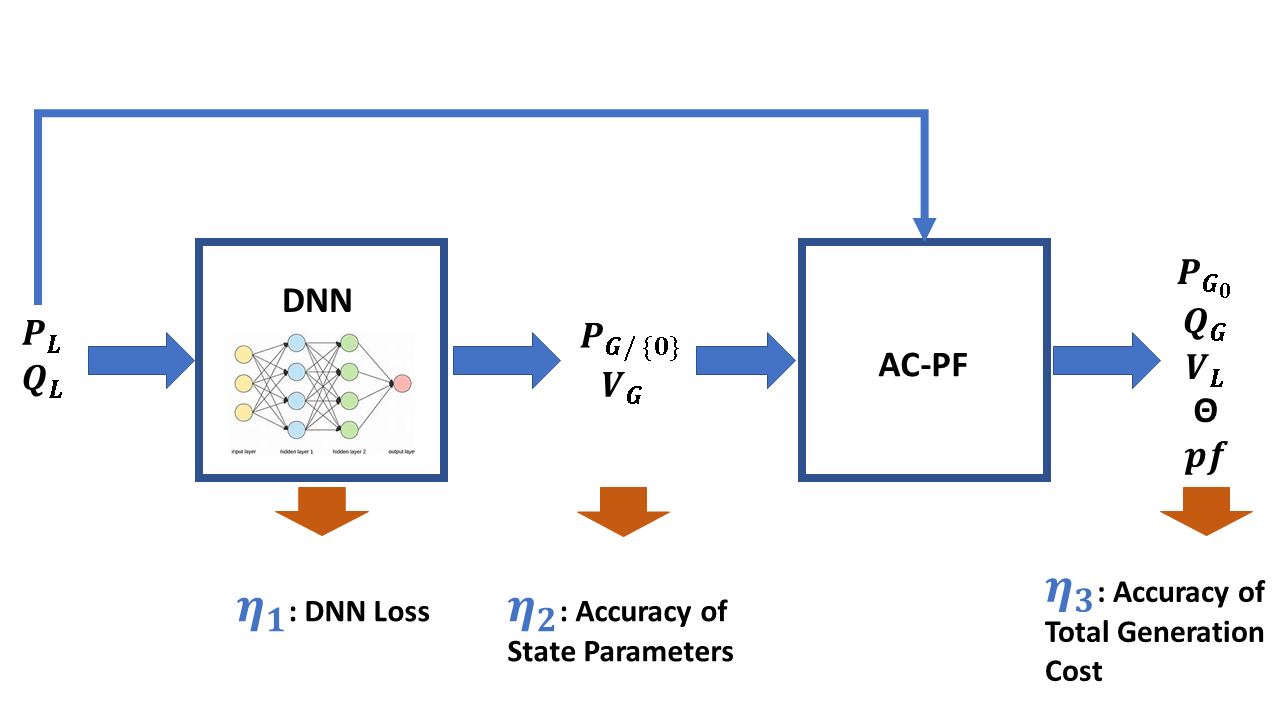}
	\caption{The online operation framework of DNN based OPF predictor}
	\label{fig:DNN-OPF}
\end{figure}

We now summarize the approaches proposed by existing works that use DNNs for the OPF problem \cite{Sun2018, Canyasse2017, PanDeepOPF2019, Zamzam2020}. Fig.~\ref{fig:DNN-OPF} shows an illustration of the overall methodology. The goal of the DNN is to approximate the non-linear mapping between the system load and the OPF solution. 
Let $\hv(\xv_k^{(m)},\wv)$ denote a parametric function, specifically a DNN under topology $m$, in our case, that takes the system load as inputs and produces the OPF outputs. Herein, $\wv$ denotes the parameters of the DNN. Further, let $\mathcal{T}_m = \{ \xv_{k}^{(m)}, \yv^{(m)}_{k} \}^{K_m}_{k = 1}$ denote the input-output pair for the OPF problem under configuration $m$. Herein, $K_m$ denotes the number of training samples and subscript $k$ denotes the training sample's index. 
For the AC OPF problem, 
the inputs correspond to the real and reactive power demand at each nodes, i.e., $\xv_{k}^{(m)} = [\pv_{D,k}^{(m)};\qv_{D,k}^{(m)}]$, 
where  $\pv_{D,k}^{(m)} \ \text{and} \ \qv_{D,k}^{(m)}$ are the vector of real/reactive power demands, i.e., $\pv_{D,k}^{(m)} = [P_{D_i,k}^{(m)}]_{i \in \mathcal{N}}$ (and $\qv_{D,k}^{(m)}$ follows a similar definition). The output corresponds to the real power generation dispatch and the generation voltages, i.e., $\yv^{(m)}_{k} = [\pv^{(m)}_{G,k};\vv^{(m)}_{G,k}]$ obtained by solving the AC OPF problem. Herein, $\pv^{(m)}_{G,k} = [P^{(m)}_{G_i,k}]_{i \in \mathcal{G} \setminus \{0 \}}$ consists of a vector of power generation at all buses except the slack bus (note that the generation at the slack bus can be determined by solving the AC power flow problem with the other generations specified) and $\vv^{(m)}_{G,k} = [V^{(m)}_{i,k}]_{i \in \mathcal{G}}$. The parameters of the DNN under topology $m$ are trained to minimize an empirical cost function given by
\begin{align}
    J_{\mathcal{T}_{m}} (\wv) = \frac{1}{K_m} \sum_{k \in  \mathcal{T}_m} || \yv^{(m)}_k - \hv(\xv_k^{(m)},\wv) ||^2. \label{eqn:loss_ML}
\end{align}
The cost function minimizes the difference between the actual OPF outputs and those predicted by the DNN. Following offline training, the DNN is deployed online to predict the generation outputs for given load inputs. We note that once $[\pv^{(m)}_{G,k};\vv^{(m)}_{G,k}]$ are predicted by a trained DNN, the other system parameters (such as the nodal voltages/power injections, etc. at the non-generator buses) can be recovered by solving AC power flow problem as shown in Fig.~\ref{fig:DNN-OPF}. Note that solving the AC power flow problem is computationally extremely fast as compared to solving the AC OPF problem, and hence, adds only a small computational overhead on the DNN approach \cite{Zamzam2020}.

\emph{Drawbacks of Existing Work:}  The main drawback of existing works is that the DNN predictions remain effective only as long as the topology of the system remains fixed. As noted before, topology reconfigurations are increasingly being adopted in power grids to ensure the economic operation and reliability \cite{Hedman2011, KoradTPS2013, RogersDFACTS2008}. While it is certainly possible to retrain the model when the system topology is changed, retraining from scratch will require significant amounts of training data and time. Alternatively, the system operator can train separate DNNs for each system configuration. But this would require a significant amount of computational resources. Moreover, the operator must know all possible topology reconfigurations beforehand, which is not possible, since unforeseen contingencies may arise during power system operations.

\section{A Meta Learning Approach for the OPF Problem Under Topology Reconfiguration}
To overcome these challenges, in this work, we seek to build an ML model for the OPF problem that can be rapidly adapted to a new system configuration. MTL is ideally suited to tackle this problem \cite{FinnMAML2017}. MTL is a training methodology that is suited to learn a series of related tasks; when presented with a new and related task, MTL can quickly learn this task from a small amount of training data samples. MTL algorithm consists of two phases, an offline training phase (also called the meta-training phase) and an online training phase (adaptation for the new task). During the offline training phase, MTL finds a set of a good initialization parameters for the series of related tasks. During the online phase, MTL uses the initialization parameter to quickly adapt the model parameters to a new task using a few gradient updates with a few training samples.

\subsection{MTL Description}
We now present the details of the proposed MTL approach. As noted in Section~2.1, we consider $M$ different grid topologies. Assume that during the offline training phase, the system operator has access to OPF training data samples from $M^*<M$ topologies. 
We denote by offline training data set by $\mathcal{T}_{\text{offline training phase}}  = \{ \mathcal{T}_1,\mathcal{T}_2,\dots,\mathcal{T}_{M^*} \}.$ During the offline training phase, MTL uses $\mathcal{T}_{\text{offline training phase}}$ to find a set of parameters $\wv_{\text{MTL}}$ that minimizes the loss function given by 
\begin{align}
J_{\text{MTL}} =  \sum^{M^*}_{m = 1}  J_{\mathcal{T}_m} (\wv - \nabla J_{\mathcal{T}_m} (\wv)), \label{eqn:MTL_train}
\end{align}
where $J_{\mathcal{T}_m}$ is defined in \eqref{eqn:loss_ML}.
Thus, $\wv_{\text{MTL}} = \argmin_{\wv} J_{\text{MTL}}.$ As evident from \eqref{eqn:MTL_train}, MTL aims to find an initialization point $\wv_{\text{MTL}}$ from which a single gradient update on each topology in $\{ 1,2,\dots,M^*\}$ yields minimal loss on that topology. Since the OPF prediction task under different topologies are related, if we succeed to find a good initialization point for the tasks in $\{ 1,2,\dots,M^*\},$ we can expect this point to be a good initialization point for any topology. Finn et. al. \cite{FinnMAML2017} proposed a gradient based method to solve the optimization problem \eqref{eqn:MTL_train}, which we summarize in Algorithm~1.

\begin{algorithm}[H]
	\caption{Offline Training for MTL}
    \textbf{Input:} Training dataset $\mathcal{T}_{\text{offline training phase}}$, Step sizes $\alpha$,$\beta$\\
    \textbf{Output:} $\wv_{\text{MTL}}$: Optimal meta parameter
	\begin{algorithmic}[1]
	\STATE Randomly initialize $\wv_{\text{MTL}}$
		\WHILE{not done}
		      \STATE Sample batch of tasks $\mathcal{T}_m \in \mathcal{T}_{\text{offline training phase}}$
		      \FORALL{$\mathcal{T}_m$}
		             \STATE Evaluate $\nabla J_{\mathcal{T}_m} (\wv_{\text{MTL}}) $ using $\mathcal{T}_m$
		             \STATE Compute adapted task model parameters with gradient descent: $\wv^{\prime}_m  = \wv_{\text{MTL}} -\beta \nabla J_{\mathcal{T}_m} (\wv_{\text{MTL}}) $
		      \ENDFOR
		      \STATE Update $\wv_{\text{MTL}} \leftarrow \wv_{\text{MTL}} -\alpha \nabla \sum^M_{m = 1}  J_{\mathcal{T}_m} (\wv^{\prime}_m) $
		\ENDWHILE
		\STATE Return $\wv_{\text{MTL}}$
	\end{algorithmic}
\end{algorithm}
In Algorithm~1, $\wv_{\text{MTL}}$ are the meta-weights (i.e., the initialization weights) for the related tasks, and  $\wv^{\prime}_m$ are the task-specific weights for the training topology $m$ (obtained from a single gradient update on $\wv_{\text{MTL}}$). The notation $\nabla J_{\mathcal{T}_m} (\wv) $ denotes the gradient of the loss function (defined in \eqref{eqn:loss_ML} computed using the dataset $\mathcal{T}_m$) with respect and weights $\wv.$ Finally $\alpha$ and $\beta$ denote the step sizes for the gradient updates.

During the online training process, assume that the system operator changes the power system topology to a new configuration that does not belong to the dataset in offline training phase. Let $\mathcal{T}^{(new)}\notin \mathcal{T}_{\text{offline training phase}}$ denote the training dataset from the new system configuration. Note that $\mathcal{T}^{(new)}$ may consist of only a few data points as compared to the offline training data. MTL finds the task-specific parameters for this new topology by performing gradient update as 
$$\wv_{new}  =   \wv_{\text{MTL}} -\gamma \nabla J_{\mathcal{T}^{(new)}} (\wv) .$$ The overall procedure for OPF using the MTL approach is presented in Algorithm~2.
\begin{algorithm}[H]
	\caption{MTL Procedure}
    \textbf{Input:} $\wv_{\text{MTL}}$, $\mathcal{T}^{(new)}$,$\gamma$\\
    \textbf{Output:} $\wv_{new}$: Adapted parameters for new configuration\\
	\begin{algorithmic}[1]
		\WHILE{system in operation}
		      \STATE Change system to new configuration
		      \STATE Obtain training samples from the dataset of new configuration $\mathcal{T}^{(new)}$
		             \STATE Compute the adapted parameters with gradient descent: $\wv_{new}  = \wv_{\text{MTL}} -\gamma \nabla J_{\mathcal{T}^{(new)}} (\wv) $
		      
		\ENDWHILE
	\end{algorithmic}
\end{algorithm}

\begin{figure}[!t]
	\centering
	\includegraphics[width=0.4\textwidth]{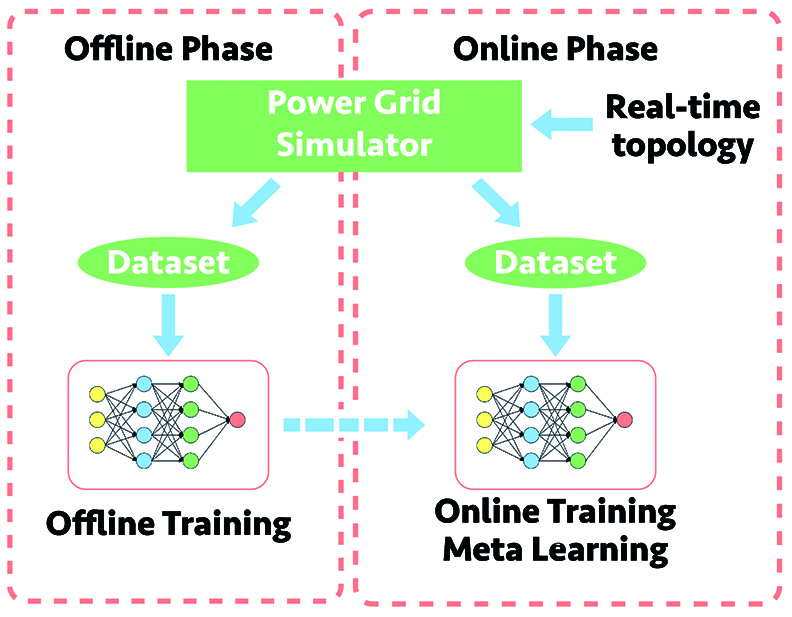}
	\caption{Schematic diagram of MTL implementation for OPF.}
	\label{fig:MTL_Implementation}
\end{figure}

\subsection{Implementation}
A schematic diagram illustrating the proposed MTL implementation is shown in Fig.~\ref{fig:MTL_Implementation}. In the offline phase, the system operator uses a power grid simulator to generate the training data set $\mathcal{T}_{\text{offline training phase}}$. The data is subsequently used to train a DNN as in Algorithm~1. During real-time operation, assume that the system operator plans a topology reconfiguration. During the online training phase, the system operator takes the new system configuration as input to a power grid simulator and generates a few new data samples for the online training phase. Then, the new samples are used to quickly fine-tune the machine learning model as in Algorithm~2. Following retraining, the new model can be used to predict the generator outputs. The online training procedure must be repeated once the system topology is changed.

\subsection{Ensuring Feasibility}
The OPF solution predicted by the DNN is feasible when they satisfy the active power generation/ nodal voltage limits, which are specified in \eqref{eqn:constrain_P}, \eqref{eqn:constrain_Q}, \eqref{eqn:constrain_V}. In order to ensure the feasibility of DNN proposed solution, we take the following approach proposed in \cite{PanXiang2021,Zamzam2020}. First, we perform a linear transformation for the active power generation/ nodal voltage magnitudes as follows: 
\begin{align}
    P_{G_i}(\rho_i) = \rho_i (P^{\max}_{G_i}-P^{\min}_{G_i})+P^{\min}_{G_i},
    \rho_i \in [0,1], 
    i \in \mathcal{G} \setminus \{0\}
    \label{eqn:linear scaling of P},
\end{align}

\begin{align}
    V_{i}(\sigma_i) = \sigma_i (V^{\max}_{i}-V^{\min}_{i})+V^{\min}_{i},
    \sigma_i \in [0,1], 
    i \in \mathcal{G} 
    \label{eqn:linear scaling of V}.
\end{align}
Then, rather than directly predicting the $P_{G_i}, i \in \mathcal{G} \setminus \{0\}$ and $V_{i} , i \in \mathcal{G}$ from the DNN, we predict the scaling parameters $\rho_i$ and $\sigma_i.$ To this end, we apply the sigmoid activation function at the output layer of the DNN to ensure that the predicted values of $\rho_i$ and $\sigma_i$ are restricted between the range $[0,1]$. By doing so, the reconstructed $P_{G_i}$ and $V_{G_i}$ will not violate the limits.

Note that while the aforementioned transformation ensures the feasibility of the variables directly predicted by the DNN, i.e., $P_{G_i}, i \in \mathcal{G} \setminus \{0\}$ and $V_{i} , i \in \mathcal{G},$ it does not ensure 
that feasibility of all the system variables -- specifically, those recovered by solving the AC power flow problem (recall Fig.~\ref{fig:DNN-OPF}).
For this reason, we calibrate the voltage constraints while generating the training dataset to avoid such violations \cite{Zamzam2020}. Specifically, in topology $m$ we calibrate the voltage constraints as 
\begin{align}
    V^{\min}_i - \lambda \leq V^{(m)}_{i} \leq V^{\max}_i+\lambda, \forall i \in \mathcal{N}
    \label{eqn:V calibration},
\end{align}
where $\lambda$ is a calibration parameter that is set to a small value. This calibration ensures that the DNN is trained to predict voltage magnitudes that lie strictly in the interior of feasible region, and hence mitigates the infeasibility caused by the approximation errors of the DNN. Finally, one can also ensure the feasbility of reactive power generations using a similar procedure. We omit and details here and refer the reader to \cite{Zamzam2020}.

\section{Simulations}
In this section, we verify the effectiveness of the proposed MTL approach using simulations and present the results.

\subsection{Algorithms and Metrics}
Under MTL, the offline and online training are performed according to Algorithm~1 and 2.
We compare the performance of MTL against three other training methods, namely, ``learn from scratch'' and ``learn from a joint training model'' and ``learn from closet model''.

\begin{itemize}
    \item In ``Learn from scratch'', there is no pretraining. During the online phase, following topology reconfiguration, a DNN's weights are intialized to random values, and trained using the OPF dataset from the new topology.
    \item The ``Learn from joint training model'' is described in Algorithm~3. During the offline training phase, a DNN is trained using the dataset $\mathcal{T_{\text{offline training phase}}},$ which combines the training data from topologies $1,\dots,M^*$. During the online phase, following topology reconfiguration, the DNN's weights are fine-tuned (from the pre-trained values) using OPF data from the new topology, similar to the MTL online training phase. 
    \item The ``learn from closet model'' is described in Algorithms~5 and 6. During the offline training phase, we train a separate DNN for each topology $1,\dots,M^*$. During the online phase, we choose the DNN that achieves the best prediction performance on the new topology at hand (step 4 of Algorithm~6). Then, we fine-tune its weights using OPF data from the new topology.
\end{itemize}
We henceforth refer to ``Learn from joint training model'' and ``Learn from the closest model''  as ``Pretrain1'' and ``Pretrain2'' respectively.

\begin{algorithm}[htb]
	\caption{Offline Training for pretrain1}
    \textbf{Input:} $\mathcal{T}_{\text{offline training phase}},\alpha$\\
    \textbf{Output:} $\wv_{\text{pretrain1}}$: The initial parameters (model) that developed based on joint training\\
	\begin{algorithmic}[1]
		\WHILE{not done}
		      \STATE Update $\wv \leftarrow \wv -\alpha \nabla J_{\mathcal{T_{\text{offline training phase}}}} (\wv) $
		\ENDWHILE
	\end{algorithmic}
\end{algorithm}

\begin{algorithm}[htb]
	\caption{Offline Training for pretrain2}
    \textbf{Input:} $\mathcal{T}_{\text{offline training phase}},\alpha$\\
    \textbf{Output:} $W_{pretrain2}$ = \{$\wv_1,\wv_2,\ldots,\wv_{M^*}$\}: The set of parameters (model) for each task in offline training phase $\mathcal{T}_{\text{offline training phase}}$\\
	\begin{algorithmic}[1]
		\FORALL {$\mathcal{T}^{(m)} \in \mathcal{T}_{\text{offline training phase}}$}
		     \WHILE{not done}
		             \STATE Update $\wv_m \leftarrow \wv_m -\alpha \nabla J_{\mathcal{T_{\text{m}}}} (\wv_m) $
		     \ENDWHILE\\
		     $W_{pretrain2} \ \stackrel{append}{\Leftarrow} \   \wv_m$
		\ENDFOR
	\end{algorithmic}
\end{algorithm}

\begin{algorithm}[htb]
	\caption{Online training for pretrain2}
    \textbf{Input:} $W_{pretrain2}$, $\mathcal{T}^{(new)}$,$\gamma$\\
    \textbf{Output:} $\wv_{new}$: Adapted parameters for new configuration\\
	\begin{algorithmic}[1]
		\WHILE{system in operation}
		      \STATE Change system to new configuration
		      \STATE Obtain training samples from the dataset of new configuration $\mathcal{T}^{(new)}$
		      \STATE Find the model that performs best on new task: $\wv_{\text{best}} = \mathop{\arg\min}_{m} \ J_{\mathcal{T}^{(new)}} (\wv_m)$ 
		      \STATE Compute the adapted parameters with gradient descent: $\wv_{new}  = \wv_{\text{best}} -\gamma \nabla J_{\mathcal{T}^{(new)}} (\wv) $
		      
		\ENDWHILE
	\end{algorithmic}
\end{algorithm}

The online operation framework of the two-step DNN based OPF solver is presented in Fig~\ref{fig:DNN-OPF} (used for the testing data). In the first step, given the active and reactive power demand at the load buses, the trained DNN predicts the active power generations (except that on the slack bus) and the voltage magnitudes at the generator buses. Then, all other system state parameters (e.g. $ P_{\mathcal{G}_0}, Q_{\mathcal{G}}, V_L, \theta$ and branch power flow \emph{pf}) can be reconstructed by solving simple AC power flow equations.

The performance of the DNN based OPF solver is assessed by three metrics, indicated in Fig.~\ref{fig:DNN-OPF}. The first metric $\eta_1$ is the DNN validation loss, which is defined in \eqref{eqn:loss_ML}. The second metric $\eta_2$ is the accuracy of the state parameters, defined in \eqref{eqn:Metric2}, where $2\vert \mathcal{G} \vert-1$ is the dimension of DNN output, $\hat{y}^{(m)}_{k,d}$ is the predicted state parameter and $y^{(m)}_{k,d}$ is the corresponding real value. The third metric $\eta_3$ is based on the accuracy of total generation cost. The total generation cost is defined as $cost = \sum_{i \in \mathcal{G}} C_i (P^{(m)}_{G_{i}})$, and introduced in \eqref{eqn:OPF_AC}. The $\eta_3$ is defined in \eqref{eqn:Metric3}, where $\hat{cost}^{(m)}_{k}$ is the predicted total generation cost and ${cost}^{(m)}_{k}$ is the corresponding real value.

\begin{equation}
\eta_2 = 1-\frac{1}{K^{(m)}}\sum_{k=1}^{K^{(m)}} \frac{1}{2\vert \mathcal{G} \vert-1} \sum_{d=1}^{2 \vert \mathcal{G} \vert-1} \Big{|}  \frac{\hat{y}^{(m)}_{k,d}-y^{(m)}_{k,d}}{y^{(m)}_{k,d}}  \Big{|}, \label{eqn:Metric2}
\end{equation}

\begin{equation}
\eta_3 = 1-\frac{1}{K^{(m)}}\sum_{k=1}^{K^{(m)}} \Big{|}  \frac{\hat{cost}^{(m)}_k-{cost}^{(m)}_k}{{cost}^{(m)}_k}  \Big{|}, \label{eqn:Metric3}
\end{equation}

\subsection{Data Creation and DNN Settings}
The power system models are built using the MATPOWER simulator \cite{Matpower2011}. The training and testing data are generated using MATPOWER's AC OPF solver. 
We test the algorithms using the IEEE-14, 30 and 118 bus systems. For each bus system, we generate $M = 100$ different grid topologies, where each topology is obtained by randomly disconnecting a set of transmission lines and changing the line impedances randomly within $30 \%$ of their original values. For each topology, we create a set of $1000$ data points, where each data point corresponds to a different load value obtained by adding a random load perturbation to the base values (that are obtained by the MATPOWER simulator). The maximum load perturbation is restricted to $70 \%$ of the original values. We consider the quadratic OPF cost, and use the default generation cost values in MATPOWER. Changes to the system topology will lead to changes in the power flows, leading to a different OPF solution. We exclude the system topologies which produce an infeasible solution from our dataset. In our simulations, we allocate $M^* = 70$ tasks to the offline training phase, and the rest $30$ tasks (denoted as new tasks) to the online training phase.

We implement the neural network model and the MTL training using PyTorch. 
The DNN settings for different bus systems used in our simulations are enlisted in Table~\ref{tbl:NN_arch}. We use the ReLu activation function at the hidden layers, and the Sigmoid activation function at the output layers. 

In the offline training process, for each pretraining method, we use the ``Adam'' optimizer with a learning rate of $0.001$ and use $1000$ training epochs. 
The L2 regularization is applied to prevent over-fitting, and weight decay is $0.001$. For the online training phase, unless specified otherwise, we use $50$ training samples during for fine-tuning the weights. Further, we use the stochastic gradient descent (SGD) optimizier with a learning rate is $0.1$, and the weight decay is $0.001$.

\begin{table}[!t]
	\centering
    	\begin{tabular}{|p{30pt}<{\centering}|p{40pt}<{\centering}|p{40pt}<{\centering}|p{40pt}<{\centering}|p{40pt}<{\centering}|}
		\toprule
		\hline
		Case & Neurons of input layer & Number of hidden layer & Neurons per hidden layer & Neurons of output layer \\
		\hline
		Case14 & 22 & 3 & 64/32/16 & 9 \\
		\hline
		Case30 & 40 & 3 & 128/64/32 & 11 \\
		\hline
		Case118 & 198 & 3 & 256/128/64 & 107 \\
		\hline
		\bottomrule
	\end{tabular}
	\caption{The neural network setting for each test case}
	\label{tbl:NN_arch}
\end{table}

\subsection{Results}
\begin{figure*}[!t]
	\centering
	\begin{subfigure}{0.32\textwidth}
	\includegraphics[width=1\textwidth]{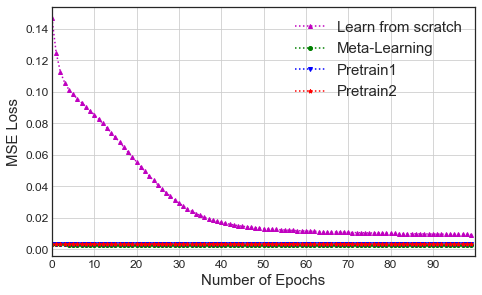}
	\caption{$\eta_1$: DNN Loss}
	\end{subfigure}
	\begin{subfigure}{0.32\textwidth}
	\includegraphics[width=1\textwidth]{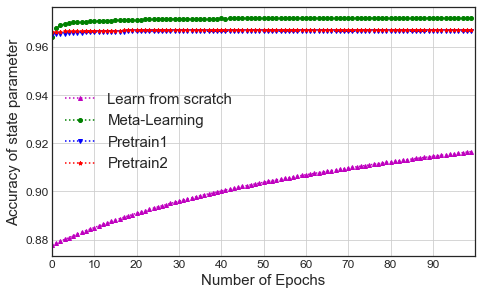}
	\caption{$\eta_2$: Accuracy of State Parameter}
	\end{subfigure}
	\begin{subfigure}{0.32\textwidth}
	\includegraphics[width=1\textwidth]{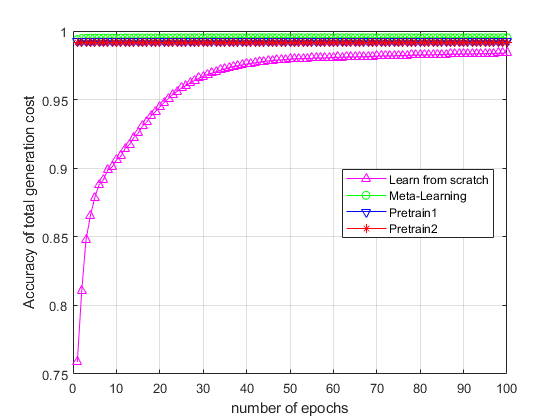}
	\caption{$\eta_3$: Accuracy of Total Generation Cost}
	\end{subfigure}
	\caption{Visualization of online training progress based on 50 training samples from the new task. Comparison of MTL with other benchmarks using the different metrics for IEEE-118 bus system.}
	\label{fig:118M}
\end{figure*}

The simulation results are presented in Fig.~\ref{fig:118M},\ref{fig:14M},\ref{fig:30M} and Tables~\ref{tbl:Comparison_detail}, \ref{tbl:Feasibility}. 
For brevity, we only present the results from the IEEE~118 bus system in Fig.~\ref{fig:118M}. The results from the IEEE-14 and 30 bus systems are relegated to the Appendix. The results in all the bus systems follow a similar trend. 

Fig.~\ref{fig:118M} and Table~\ref{tbl:Comparison_detail} present the accuracy results based on the different metrics defined in Section~4.1. It can be observed that MTL achieves a very high prediction accuracy of over $97 \%$ ($\eta_2$) and over $99 \%$ ($\eta_3$) with less than $10$ training epochs. This shows that MTL can rapidly adapt to the new system configuration starting from the initialization point $\wv_{\text{MTL}}.$ In contrast, training from scratch from a random initialization takes a significantly greater number of gradient updates.

Furthermore, MTL also achieves the highest accuracy as compared to the other pretraining methods (Pretrain~1 and 2) and lower loss. More importantly, we also observe that online training with a very number of data samples (i.e., $50$ OPF data samples from the new topology in our case) does not significantly improve the performance of other pretraining methods as observed in Table~\ref{tbl:Comparison_detail} (sometimes, we also observed that for other pretraining methods, online training with only a few data samples may result in worse performance due to over-fitting). Thus, with the other pretraining methods, the accuracy is limited to the performance achieved during the offline training phase.

\begin{figure}[!t]
	\centering
	\includegraphics[width=0.4\textwidth]{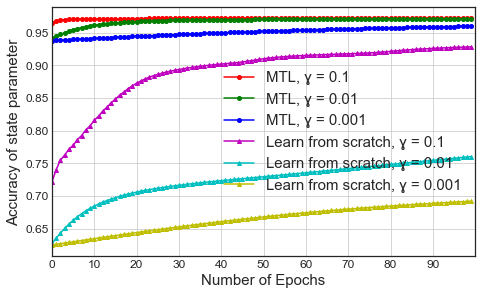}
	\caption{MTL/ Learn from scratch, online training performance under different learning rates $\gamma.$}
	\label{fig:MTL_LR}
\end{figure}

From Fig.~3(b), we observe that for MTL, most of the performance improvement occurs within the first few epochs. Thus, MTL is suitable for online training with a very few data samples and a very few training epochs. 
A comparison of the performance of MTL and learn from scratch for different training rates during the online training phase is presented in Fig.~\ref{fig:MTL_LR}, which once again shows that MTL can train quickly. 
We also present the results for feasibility of the predicted OPF solution in Table~\ref{tbl:Feasibility}. The feasibility rate is calculated as $\emph{fr} = \frac{n_f}{n_t}$, where $n_f$ denotes the number of testing sample that achieves feasible solution, and $n_t$ denotes the total number of testing samples. The results show that the adjustments made to the training process proposed in Section~3.3 is able to ensure that MTL achieves very high feasibility rate.

Besides the advantages of MTL in terms of accuracy, another advantage is its ability to quickly produce an an initialization model (i.e., the offline training). In Table~\ref{tbl:Time_require}, we enlist the time required to produce the intialization model of MTL and other pretraining methods for different bus systems. It can be observed that MTL takes significantly less time than the other pretraining methods. Moreover, as compared to the Pretrain2 method, which requires a separate DNN to be trained and stored for each power grid topology, MTL requires a single DNN model to be stored. Thus, MTL also significantly reduces the storage burden in comparison to the Pretrain2 method.

\begin{table}[!t]
	\centering
    	\begin{tabular}{c c c c c c}
		\toprule
		\hline
		Metric & Method & Epoch 0 & Epoch 1 & Epoch 10 & Epoch100 \\
		\hline
		\multirow{3}{*}{$\eta_1$} & MTL & 0.0105 & 0.0040 & 0.0030 & 0.0028 \\

		 & pretrian1 & 0.0051 & 0.0051 & 0.0051 & 0.0051 \\
		 
		 & pretrian2 & 0.0059 & 0.0060 & 0.0060 & 0.0059 \\
        \hline
		\multirow{3}{*}{$\eta_2$} & MTL & 0.9372 & 0.9642 & 0.9707 & 0.9720 \\
		
		& pretrian1 & 0.9598 & 0.9597 & 0.9597 & 0.9598 \\
		 
		 & pretrian2 & 0.9567 & 0.9564 & 0.9565 & 0.9567 \\

        \hline
		\multirow{3}{*}{$\eta_3$} & MTL & 0.9886 & 0.9937 & 0.9948 & 0.9950 \\
		
		& pretrian1 & 0.9925 & 0.9925 & 0.9925 & 0.9925 \\
		 
		 & pretrian2 & 0.9915 & 0.9914 & 0.9915 & 0.9915 \\
		 
		\hline
		\bottomrule
\end{tabular}
\caption{The online training performance of each pretrain method}
\label{tbl:Comparison_detail}
\end{table}

\begin{table}[!t]
	\centering
    	\begin{tabular}{|c|c|c|c|}
		\toprule
		\hline
		\diagbox{Method}{Feasibility}{Case} & 14-bus & 30-bus & 118-bus\\
		\hline
		Learn from scratch & 0.978 & 0.989 & 0.991 \\
		\hline
		MTL & 0.998 & 0.994 & 0.994 \\
		\hline
		pretrain1 & 0.989 & 0.993 & 0.994 \\
		\hline
		pretrain2 & 0.989 & 0.993 & 0.994 \\
		\hline
		\bottomrule
	\end{tabular}
	\vspace{0.1 cm}
	\caption{Feasibility rate after 100 epochs for each method and test case.}
	\label{tbl:Feasibility}
\end{table}

\begin{table}[!t]
	\centering
    	\begin{tabular}{|c|c|c|c|}
		\toprule
		\hline
		Pretraining method & 14-bus & 30-bus & 118-bus\\
		\hline
		MTL & 3min 55sec & 3min 40sec & 9min 6sec \\
		\hline
		pretrain1 & 11min 34sec & 13min 29sec & 104min 31sec \\
		\hline
		pretrain2 & 24min 35sec & 28min 38sec & 151min 16sec \\
		\hline
		\bottomrule
	\end{tabular}
	\vspace{0.1 cm}
	\caption{Computational time for the pretraining methods during the offline training phase.}
	\label{tbl:Time_require}
\end{table}

\begin{figure}[!t]
	\centering
	\includegraphics[width=0.4\textwidth]{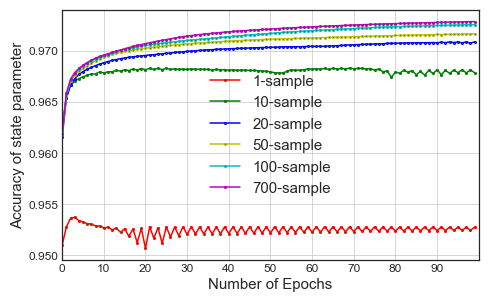}
	\caption{Test case under IEEE-118 bus system, the online training performance of MTL based on \{1, 10, 20, 50, 100, 700\} samples with 70 tasks in offline training phase. The assessment is based on $\eta_2$: accuracy of state parameter. Each model is updated according to 'SGD' optimization.The learning rate is 0.1 and weight decay is 0.001.}
	\label{fig:Sample number}
\end{figure}

\begin{figure}[!t]
	\centering
	\includegraphics[width=0.4\textwidth]{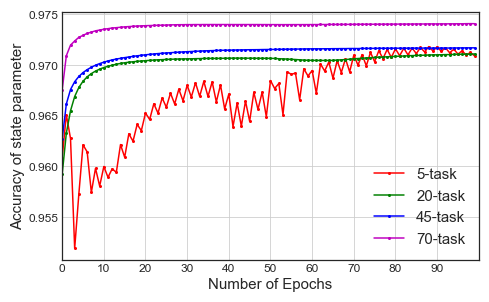}
	\caption{Test case under IEEE-118 bus system, with \{$5, 20, 45, 70$\} tasks in offline training phase and training based on 50 samples from new task. The assessment is based on $\eta_2$: accuracy of state parameter. Each model is updated according to 'SGD' optimization.The learning rate is 0.1 and weight decay is 0.001.}
	\label{fig:Task number}
\end{figure}

\begin{figure*}[!t]
	\centering
	\begin{subfigure}{0.32\textwidth}
	\includegraphics[width=1\textwidth]{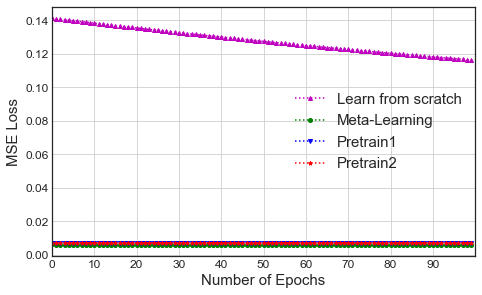}
	\caption{$\eta_1$: DNN Loss}
	\end{subfigure}
	\begin{subfigure}{0.32\textwidth}
	\includegraphics[width=1\textwidth]{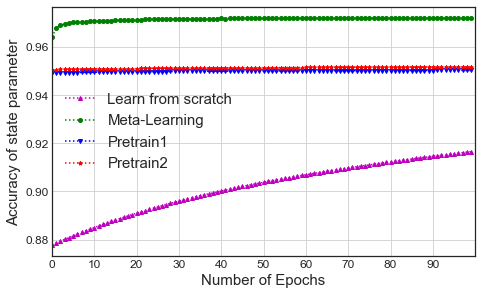}
	\caption{$\eta_2$: Accuracy of State Parameter}
	\end{subfigure}
	\begin{subfigure}{0.32\textwidth}
	\includegraphics[width=1\textwidth]{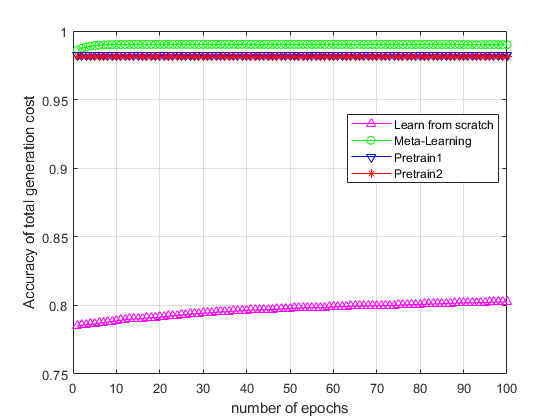}
	\caption{$\eta_3$: Accuracy of Total Generation Cost}
	\end{subfigure}
	\caption{Visualization of online training progress based on 50 training samples from the new task. Comparison of MTL with other benchmarks using the different metrics for IEEE-14 bus system. Learning rate = 0.001, Weight decay = 0.001.}
	\label{fig:14M}
\end{figure*}

\begin{figure*}[!t]
	\centering
	\begin{subfigure}{0.32\textwidth}
	\includegraphics[width=1\textwidth]{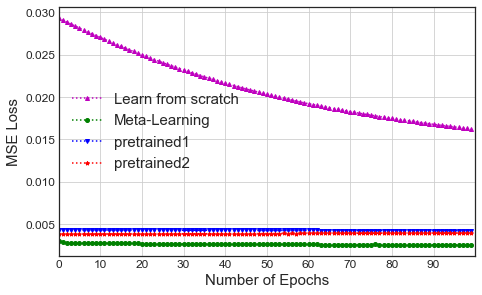}
	\caption{$\eta_1$: DNN Loss}
	\end{subfigure}
	\begin{subfigure}{0.32\textwidth}
	\includegraphics[width=1\textwidth]{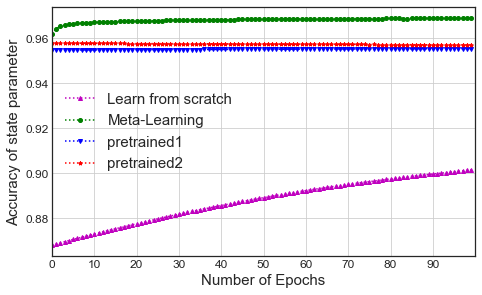}
	\caption{$\eta_2$: Accuracy of State Parameter}
	\end{subfigure}
	\begin{subfigure}{0.32\textwidth}
	\includegraphics[width=1\textwidth]{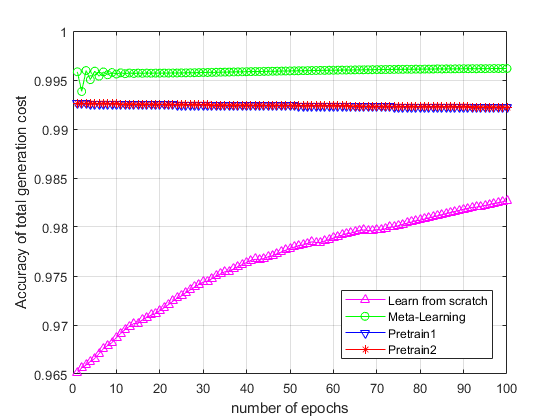}
	\caption{$\eta_3$: Accuracy of Total Generation Cost}
	\end{subfigure}
	\caption{Visualization of online training progress based on 50 training samples from the new task. Comparison of MTL with other benchmarks using the different metrics for IEEE-30 bus system. Learning rate = 0.01, Weight decay = 0.001.}
	\label{fig:30M}
\end{figure*}

Finally, we investigate the performance of MTL as a function of the online/offline training parameters. To this end, first, we investigate the the prediction accuracy (measured according to the metric $\eta_2$) as a function of the number of training samples used in the online training progress and present the results in Fig.~\ref{fig:Sample number}. We observe that MTL achieves good accuracy by fine-tuning with only $50-100$ online training samples. Increasing the number of online training samples to $700$ achieves a negligible improvement in the accuracy. This implies that MTL is good for fine tuning with a very few number of data samples, making it particularly attractive for online training. Secondly, we investigate the the MTL prediction accuracy as a function of the number of topologies used in the offline training phase $\mathcal{T}_{\text{offline training phase}}$. The results plotted in Fig.~\ref{fig:Task number} indicate that the prediction accuracy goes down when the number of topologies used in the offline training process is reduced.  This indicates that a sufficient number of toplogies are required in the offline training phase to develop an efficient MTL model.

\section{Conclusions}
\label{sec:Conc}
In this work, we have proposed a DNN based approach to the OPF problem that is trained using a novel MTL approach. The proposed approach is particularly relevant for computing OPF generation dispatch decisions under power grid topology reconfigurations. The MTL approach finds good initialization points from which the DNNs can be quickly trained to produce accurate predictions for different system configurations. Simulation results show that the proposed approach can significantly enhance the training speed and achieve better prediction accuracy as well as feasible results compared to several other pretraining methods. To the best of our knowledge, this work is the first to adopt an MTL approach in a power grid context.

\section*{Appendix: Simulation Results for IEEE-14 and 30 Bus Systems}

\bibliographystyle{IEEEtran}
\bibliography{IEEEabrv,bibliography}


\end{document}